\pdfoutput=1

\documentclass[11pt]{article}

\usepackage{acl}

\usepackage{times}  
\usepackage{latexsym}

\usepackage[T1]{fontenc}

\usepackage[utf8]{inputenc}

\usepackage{microtype}

\usepackage{inconsolata}

\usepackage{amsmath}
\usepackage{makecell}
\usepackage{multirow}
\usepackage[cjk]{kotex}
\usepackage{array}
\usepackage{booktabs}
\usepackage{graphicx}
\usepackage{array}
\usepackage[framemethod=TikZ]{mdframed}
\usepackage{xcolor}

\definecolor{cadmiumgreen}{rgb}{0.0, 0.42, 0.24}

\title{Translation of Multifaceted Data \\
without Re-Training of Machine Translation Systems}

\author{Hyeonseok Moon$^{1}$, Seungyoon Lee$^{1}$, Seongtae Hong$^{1}$ \\ {\bf \large Seungjun Lee$^{1}$, Chanjun Park$^{2}$, Heuiseok Lim$^{1\dagger}$}\\
  $^1$Department of Computer Science and Engineering, Korea University\\
  $^2$Upstage\\
  $^{1}$\texttt{\{glee889,dltmddbs100,ghdchlwls123,dzzy6505,limhseok\}@korea.ac.kr} \\ \texttt{$^{2}$chanjun.park@upstage.ai}}

\begin{document}
\maketitle
\begin{abstract}

Translating major language resources to build minor language resources becomes a widely-used approach. Particularly in translating complex data points composed of multiple components, it is common to translate each component separately. However, we argue that this practice often overlooks the interrelation between components within the same data point. To address this limitation, we propose a novel MT pipeline that considers the intra-data relation\footnote{Note that a single data point is composed of multiple components. In this sense, we use the term ``interrelation among data components'' as the same meaning as the ``intra-data relation within a data point''} in implementing MT for training data. In our MT pipeline, all the components in a data point are concatenated to form a single translation sequence and subsequently reconstructed to the data components after translation. 
We introduce a Catalyst Statement (CS) to enhance the intra-data relation, and Indicator Token (IT) to assist the decomposition of a translated sequence into its respective data components. Through our approach, we have achieved a considerable improvement in translation quality itself, along with its effectiveness as training data. Compared with the conventional approach that translates each data component separately, our method yields better training data that enhances the performance of the trained model by 2.690 points for the web page ranking (WPR) task, and 0.845 for the question generation (QG) task in the XGLUE benchmark.

\end{abstract}

\section{Introduction}
\begin{figure}[t]
\centering
\includegraphics[width=\linewidth]{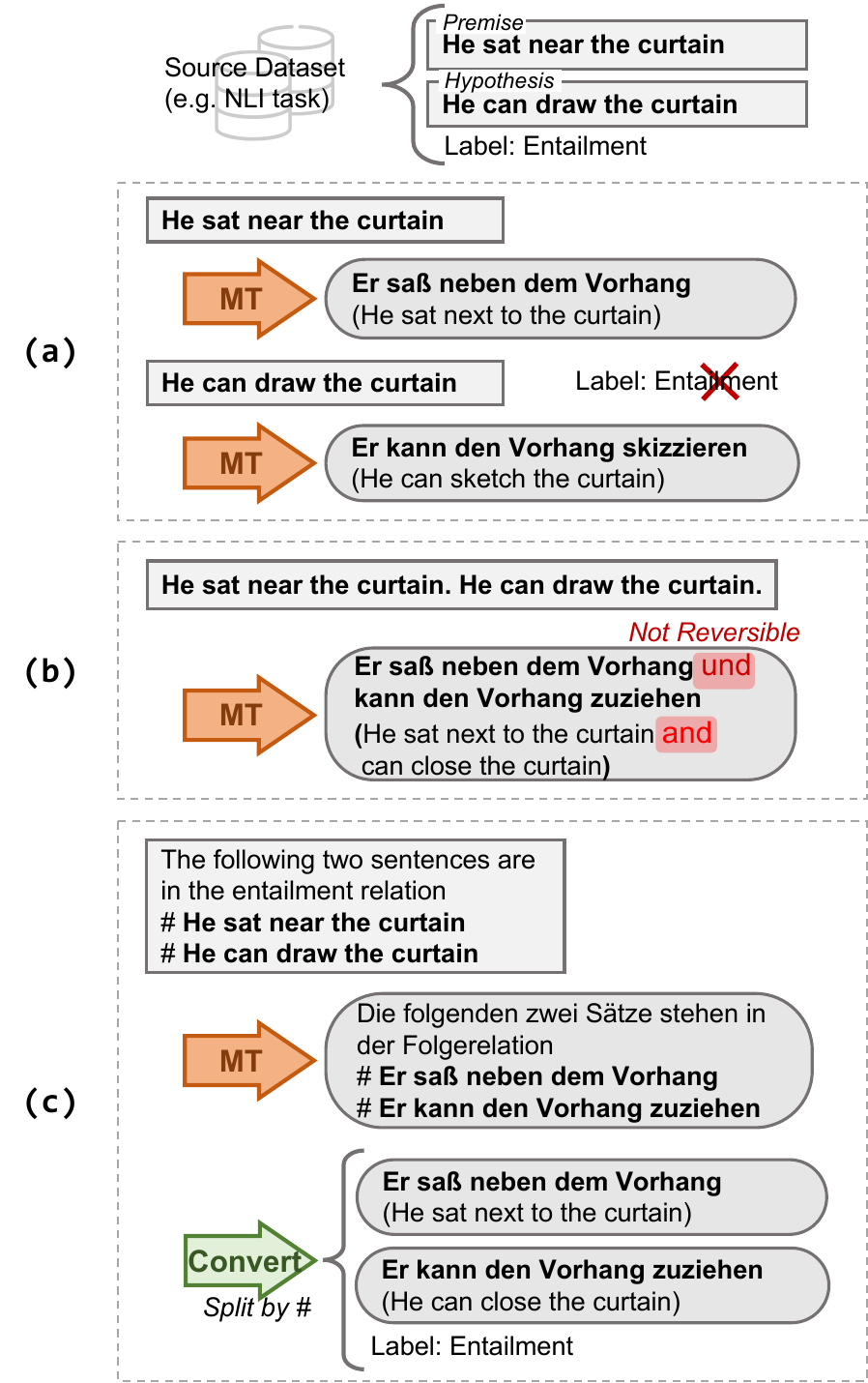}
 \caption{Example of challenges in data translation} 
 \label{fig:example}
\end{figure}

Machine translation (MT) has been developed to aid human-level utilization, with its primary focus on the accurate translation of any given sequence (\emph{i.e.}, ensuring semantic preservation and syntactic fluency) \cite{specia2020findings, guzman2019flores, martindale2019identifying, rei2020comet}. As previous MT systems have demonstrated relatively low performance \cite{daems2017identifying, vilar2006error}, their translation outputs are hardly utilized as another data source. With the ongoing advancement of MT research, the translation performance of MT systems becomes comparable to the expert human-level \cite{costa2022no, peng2023towards}, and subsequently several attempts have recently emerged to utilize MT system for the data translation process \cite{cui2023efficient, li2023bactrian, liang2020xglue, peng2023instruction, kakwani2020indicnlpsuite, chen2022frustratingly, bassignana2023multi}. Particularly, several non-English datasets are vigorously being constructed by translating English datasets \cite{bassignana2023multi, adelani2023sib, abulkhanov2023lapca}.

In applying MT to data translation, one concern we raise is the conservation of the intra-data relation during MT process. Depending on the task composition, a single data point may comprise multiple components. For example, each data point of natural language inference (NLI) task comprises three components; namely, the hypothesis and the premise along with one label. In translating such multifaceted components, we encounter dilemmas in determining the input unit, considering MT systems generally take a singular sequence. 

In dealing with this situation, current research predominantly translates each individual data component separately \cite{turc2021revisiting, bigoulaeva2023label}.
However, we argue that such data translation approaches may not yield optimal results, as the interrelation among components in the same data can easily be disregarded. As shown in Figure~\ref{fig:example}-(a), the translated pair may not accurately maintain the original label, despite the absence of any error in their respective translations. This can further derive performance degradation of the model trained with these translated datasets, as the purpose of the task is generally represented within the interrelation among data components.

Theoretically, this issue can partially be alleviated by simply concatenating all the components in a single sequence for translation. Then in translating each component, MT system can refer the semantics of other components in the same sequence. However, in this case, MT system often merges all the components and generates an inseparable result to form a natural context. As shown in Figure~\ref{fig:example}-(b), this presents challenges in distinguishing data components from the translated sequence.

\begin{figure*}[t]
\centering
\includegraphics[width=\linewidth]{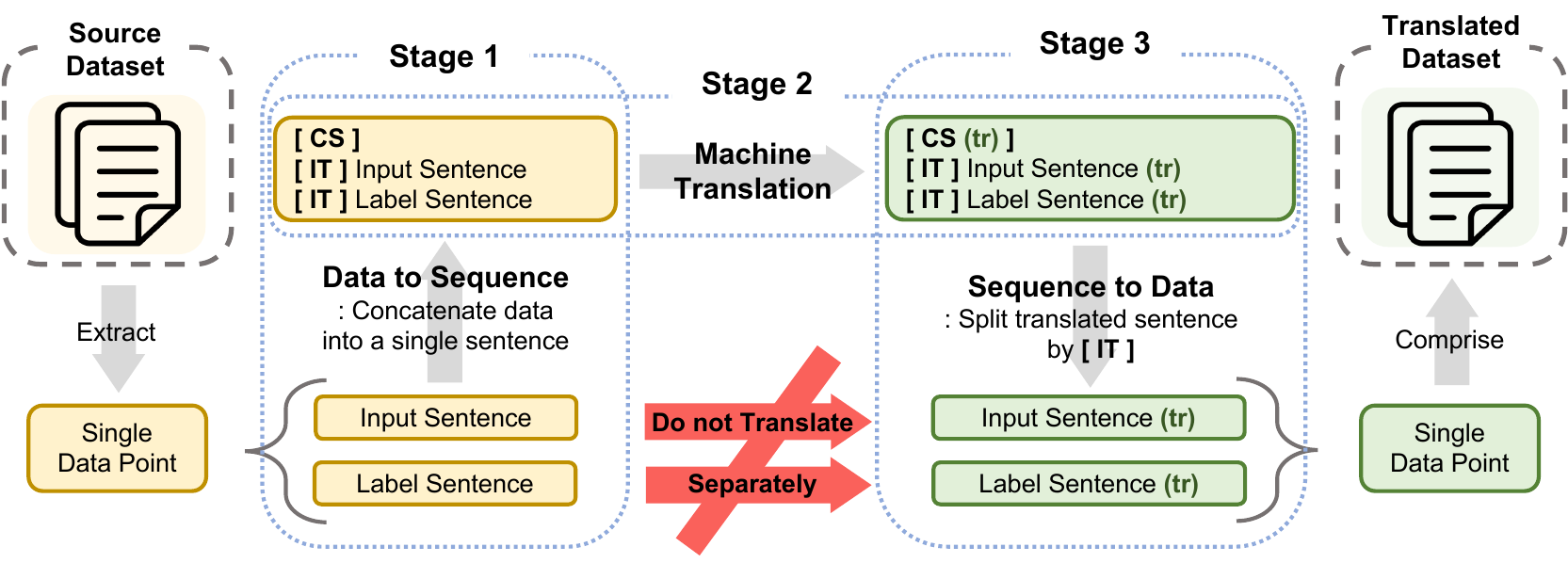}
 \caption{Relation-Aware translation pipeline. To explain the overall process, we assume data comprises two components: input sentence and label sentence. In this figure, \textcolor{cadmiumgreen}{\textbf{(tr)}} represents the corresponding translated unit.} 
 \label{fig:pipeline}
\end{figure*}

Upon these considerations, we propose a simple yet effective MT pipeline for the data translation that can be applied to any MT systems without further re-training. In particular, we propose a relation-aware translation that strategically concatenates multifaceted components into a singular sequence, as in Figure~\ref{fig:pipeline}. Especially in concatenating data components, we discern the following two aspects: \textbf{(\romannumeral 1)} the inter-relation between components should be considered in a concatenated sequence. \textbf{(\romannumeral 2)} translated sequence should be reversible (\emph{i.e.} can explicitly be converted to the translated data components). To attain these objectives, we introduce \textbf{Indicator Token (IT)} and \textbf{Catalyst Statement (CS)}. \textbf{IT} is basically designed to distinguish the location of each data component and help conversion of the translated sequence into the translated components. \textbf{CS} is devised to specify the definite relation between each component in the concatenated sequence for enhancing the inter-relation between components. Constructed sample is shown in Figure~\ref{fig:example}-(c).

For validation, we select multilingual benchmark tasks in which the maintenance of the interrelation among data components plays a critical role. Specifically, we adopt the XNLI dataset \cite{conneau2018xnli} and select two tasks in an XGLUE benchmark \cite{liang2020xglue}: Web Page Ranking (WPR) and Question Generation (QG). We construct training data for up to five languages (German, French, Chinese, Hindi, and Vietnamese) by translating the English dataset existing within each dataset. 
Subsequently, by evaluating the performance of the models trained on each translated data, we estimate the validity of each data translation strategy. Notably, our proposed data translation pipeline demonstrates a more effective strategy to attain high-quality training data, compared to the individual translation of each data component. 

\section{Related Works}
Attempts to construct training data with MT systems can broadly be divided into two major approaches. The first approach aims to construct a task-specific MT system by training with any corpus specially constructed for reaching intended goal \cite{phang2020english, ramponi2020neural, carrino2020automatic, lewis2020mlqa, duan2019zero, shen2018zero}. For instance, \citet{sowanski2023slot} trained a new translation model with a manually curated domain-specific dataset, then made a Polish training corpus for virtual assistant by translating English dataset. However, these attempts encounter difficulties in utilizing newly released assets. 

In contrast, the second approach covers attempts to use publicly released NMT models without any modification, in constructing datasets via translation \cite{mozannar2019neural, croce2019enabling, bassignana2023multi, adelani2023sib, abulkhanov2023lapca, sorokin2022ask}. Representatively, commercialized NMT systems such as DeepL~\footnote{https://www.deepl.com/translator} \cite{croce2019enabling, bassignana2023multi} or Google Translator~\footnote{https://translate.google.com/} \cite{mozannar2019neural, lee2018semi}, along with publicly released NMT models \cite{costa2022no, fan2021beyond} are adopted to construct multilingual training datasets \cite{adelani2023sib}. However, prior approaches using existing assets with modification have encountered limitations in performing accurate data translation considering the interrelation among components comprising each data. Considering these attempts and their challenges, we focus on establishing an easy-to-implement pipeline for data translation that utilizes MT systems without model modification and takes into account intra-data relation.

\section{Machine Translation for Machine}

\begin{table*}[h]
\centering
\resizebox{0.99\textwidth}{!}{
\begin{tabular}{c|l|l}
\toprule[1.5pt]
\textbf{Task} & \makecell[c]{\textbf{CS Type}} & \makecell[c]{\textbf{Catalyst Statement}} \\ \midrule
\multirow{2}{*}{\textbf{NLI}} & \textbf{Concat} & The following is a pair of sentences that are related to each other \\
{} & \textbf{Relation} & The following two sentences are in the \textbf{[LABEL]} relation \\ \midrule

\multirow{2}{*}{\textbf{WPR}} & \textbf{Concat} & The following is a group of sentences that are related to each other \\
{} & \textbf{Relation} & Using the first sentence as a query, we obtained the following search results. We evaluate these results as \textbf{[LABEL]} \\ \midrule

\multirow{2}{*}{\textbf{QG}} & \textbf{Concat} & The following is a pair of sentences that are related to each other \\
{} & \textbf{Relation} & The second sentence is a \textbf{question} that can be generated after reading the first \textbf{passage}  \\

\bottomrule[1.5pt]

\end{tabular}
}
\caption{Catalyst Statements (CSs) adopted in our experiments. Samples of constructed translation sequences are shown in Table~\ref{tab:samples}.} \label{tb:catalyt_prompt}
\end{table*}

\subsection{Problem Statement} 
In this study, we focus on potential issues of translating data constituting multiple components using conventional MT systems. Take QG task as an example, in which data constitutes passage $x$ and question $y$ as its components. We should note that there exists definite relation between these components: $x$ is a passage that can derive a question $y$, and $y$ is a question that can be retrieved from the passage $x$.

Ideally, in translating $(x, y)$ to obtain a translated pair $(x', y')$, the semantic relation between $(x, y)$ should be preserved after translation. To ensure relation-considering translation, the MT system should consider both components together even in translating respective components. This can be represented as an inference objective of maximizing probabilities displayed in Equation~(\ref{eq:ideal}).

\begin{equation} \label{eq:ideal}
    p(x'_{i} | x'_{<i}, x, y) \,\,\,,\,\,\, p(y'_{i} | y'_{<i}, x, y)
\end{equation} 

However, since MT system takes only a singular sequence, it can be challenging to impose additional constraints beyond the translating sequence. Consequently, in the conventional scenario, each component composing the same data point is individually translated instead, with an inference objective shown in Equation~(\ref{eq:current}).

\begin{equation} \label{eq:current}
    p(x'_{i} | x'_{<i}, x) \,\,\,,\,\,\, p(y'_{i} | y'_{<i}, y)
\end{equation}

In this scenario, we argue that the efficacy of translated components as training data is inevitably diminished due to the lack of consideration for the intra-data relation. Theoretically, this issue can partially be alleviated by simply concatenating two components before translation, as the MT system can simultaneously refer to the context of all components. This entails translation with an inference objective similar to Equation~(\ref{eq:concat}), where ";" denotes any form of sequence concatenation.

\begin{equation} \label{eq:concat} 
    p(z'_{i} | z'_{<i}, z) \,\,\, \text{where} \,\,\, z = [x ; y]
\end{equation}

Following the above equation, $x$ in $z$ can be translated by referring the semantics of $y$ in $z$ and vice versa. Subsequently, $x'$ and $y'$ can be yielded within the consideration of inter-relation between $x$ and $y$.

However, in this case, the translated sequence $z'$ might not be separated into $x'$ and $y'$. As the major objective of the MT system is gaining fluent context, MT systems frequently insert conjunctions between two components and merge them into inseparable semantic unit, if necessary. Then the translated sequence cannot be converted to the translated data component whether the translation is perfect or not.

In essence, the primary challenges in data translation can be encapsulated as follows:
\begin{itemize}
    \item Translating individual components hardly considers the intra-data relation within the same data point. 
    \item When these are concatenated into a single sequence without any consideration, the translated sequence might not be restored back into the respective data components.
\end{itemize}

\subsection{Our Solution: Relation-Aware Translation}
To address these issues, we propose a viable strategy for performing data translation via any conventional MT framework without any modification. 
Our strategy involves a simple three-stage pipeline as shown in Figure~\ref{fig:pipeline}. 


First, we concatenate multifaceted components into a singular sequence to enable data translation through any form of MT systems (Data to Sequence). In concatenating instances, we integrate catalyst statement (\textbf{CS}) and indicator token (\textbf{IT}) to enhance the interrelation between data components and better distinguish the location of each data component after translation. \textbf{CS} is inserted in the head of the sequence, defining the relation between data components. \textbf{IT} is attached directly in front of each data component. Samples of constructed translation sequences are shown in Table~\ref{tab:samples}, and we elaborate each role of \textbf{CS} and \textbf{IT} is elaborated in our subsequent sections.

Then we translate the concatenated sequence through the MT system. In implementing MT, we expect \text{IT} to be preserved intact after translation. If \textbf{IT} is not preserved after translation, we inevitably discard that data as we can hardly discriminate translated units for each data component. This may incur a degree of data loss; however, by conducting extensive experiments, we demonstrate that this process enables us to obtain high-quality training data from the remaining dataset.

After translation, we extract data components from the translated sequence (Sequence to Data). Specifically, we distinguish each translated component by splitting the translated sequence by the \textbf{IT}. Throughout this process, we can obtain the translated dataset, where each data point is translated with the consideration of intra-data relation. 

\subsection{Indicator Token (IT)}
In cases where two or more components constituting the data are concatenated, the most intuitive way of ensuring the sequence can be re-segmented after translation, is to accurately specify each boundary. This can be performed based on a simple punctuation (`.'). Yet a more definitive criterion is necessary as a single component can comprise multiple sentences, and punctuation can frequently be substituted to conjunctions after translation, as depicted in Figure~\ref{fig:example}. In this regard, we prepend \textbf{IT} to each data component in concatenating data into a single sequence to distinguish the location of each component after translation. Notably, we expect \textbf{IT} to remain intact after translation, thereby we can obtain a translated data point by segmenting the translated sequence by \textbf{IT}.

Representatively, we experiment with the following simple instances: \text{@}, \text{\#}, \text{*}. We take a single character form concerning any harms of semantics derived by the \textbf{IT}. We recognize that there may exist more effective instances of \textbf{IT} beyond the three examples we experimented with; we remain a room for improvement. In this paper, we focus on analyzing the impact of \textbf{IT} itself in data translation.

\subsection{Catalyst Statement (CS)}
By translating concatenated sequences, we can theoretically consider relation between the components within the data point. However, in such cases, it might be challenging to discern how these components are directly related to each other, as naive concatenation can retains semantic separation between components within the same sequence, and thereby MT systems can hardly catch their semantic relation. 

To enhance the interrelation between components within the same sequence, we propose to add an additional sentence that represents the definite relation. The purpose of its introduction is to signify the interrelated ties among the data components within the sequence to be translated, and to provide assistance by making these relations even explicit during the translation process. In essence, the aim is to substitute the task of translating seemingly semantically-separated statements with the attempt to translate a semantically-related single unit.

We denote this additional sentence as a \textbf{CS}. Particular examples we adopt in this study are shown in Table~\ref{tb:catalyt_prompt}. We define the following two types of \textbf{CS}: directly defining the relation between components (\textbf{Relation CS}) and merely serving to connect components into a single sequence (\textbf{Concat CS}). 

We use only simplified samples where other elements are excluded to objectively analyze the impact of considering intra-data relation during data translation. Specifically, these two sentences can be distinctly differentiated depending on the method of defining the relation of components. While there are potentially more possible CSs than the two we selected, we conduct experiments solely with these two representative samples to clarify our objective.

\section{Experimental Settings}

\subsection{Dataset Details}
We validate the effectiveness of our approach with the XNLI dataset \cite{conneau2018xnli} and selected two tasks in the XGLUE benchmark \cite{liang2020xglue} (WPR and QG). To acquire more general results, we conduct experiments in two to five languages for each dataset. Detailed statistics and composition of each dataset are described in Appendix~\ref{app:datastat}

\subsection{Evaluation Details}
We evaluate the validity of translation based on two primary criteria. The first is the \textbf{data reversibility}. As we have pointed out, if we translate concatenated sequence, respective components can be merged into a non-reversible element. We regard it as a translation failure, as it can hardly be utilized as training data. In estimating reversibility, we measure the percentage of the reversible data among translated sequences. 

The second criterion pertains to the \textbf{quality} of the translated data. The main objective of our MT pipeline is enhancing the value of translated data as training instances by considering intra-data relation during the translation process. To validate our goal, we evaluate the performance of the model trained on the translated data. We estimate the label accuracy for evaluating performance of NLI and WPR tasks and measure ROUGE-L \cite{lin2004rouge} for QG task. To deepen our evaluation, we compare this with the quality of the translation quality (estimated with BLEU score \cite{post2018call}) and the results from the LLM evaluation \cite{liu2023gpteval, chen2023alpagasus}.

\begin{figure*}[t]
\centering
\includegraphics[width=\linewidth]{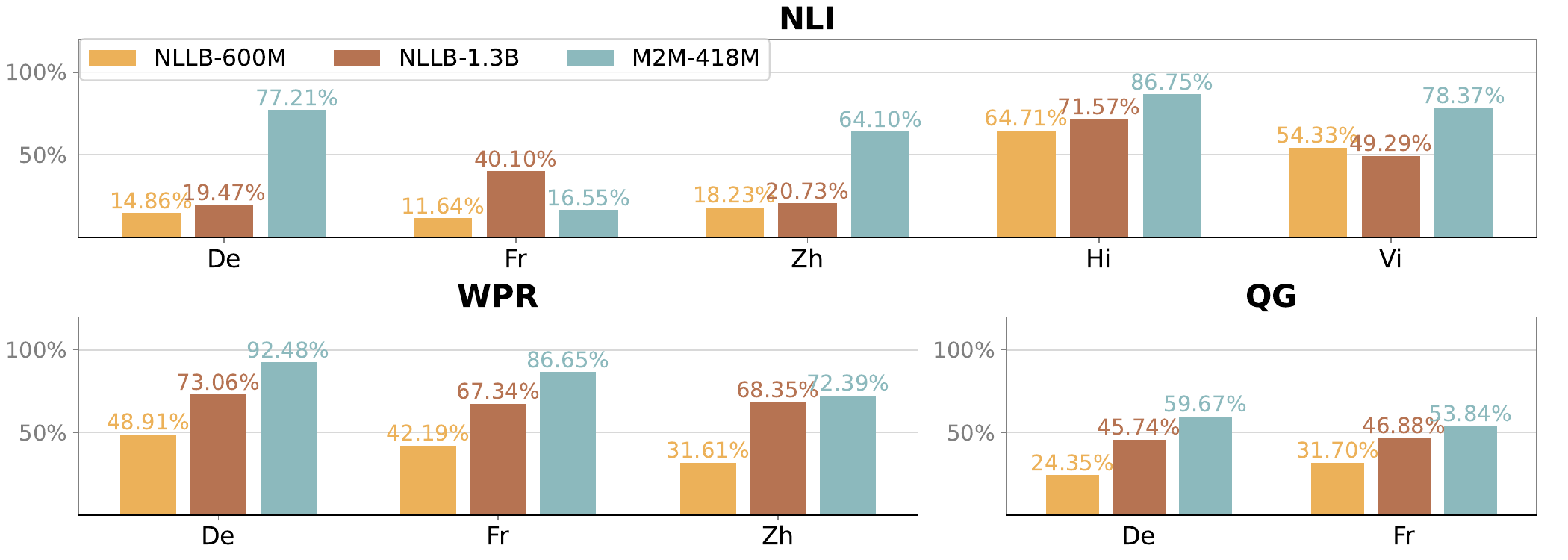}
 \caption{Data reversibility per NMT model and target dataset. For each data point, we create a single sequence by concatenating data components with a `\#' symbol and examine the preservation rate of `\#' in the translated sequence.} 
 \label{fig:problemstatement}
\end{figure*}

\subsection{Model Details}
For implementing MT, we employed the multilingual MT systems capable of processing multiple languages, NLLB \cite{costa2022no} and M2M100 \cite{conneau2020unsupervised}. Considering the verification scale and our resource constraints, we select distilled version of the original large-scale MT models: NLLB-600M, NLLB-1.3B, and M2M100-418M. After data translation, the translated data are fine-tuned with multilingual pre-trained language models to evaluate their value as training data. For NLI and WPR tasks, we adopt the XLM-R base model \cite{conneau2020unsupervised}, and for the QG task, we implement with the mT5 base model \cite{xue2021mt5}. Implementation details are included in Appendix~\ref{app:implementation}.

\section{Results and Discussion}
\subsection{Simple concatenation does not guarantee the reversibility}
In our preliminary discussions, we highlighted the issue that translating a concatenated sequence of data components may results in inseparable translated results, that cannot be converted to data components. This section provides experimental evidence supporting this claim. For each data point, we create a single sequence by concatenating data components with a `\#' symbol and examine the preservation rate of `\#' in the translated sequence. As Figure~\ref{fig:problemstatement} demonstrates, the majority of cases fail to properly maintain the integrity of the `\#'. For German training dataset in NLI task, the NLLB-1.3B model preserved only 19.47\% data points, indicating that approximately 80\% of translated sequence can not be utilized as a data component. This underscores that mere concatenation is not a viable strategy for data translation and emphasizes the need for a thoughtful approach that considers relational aspects to ensure effective data translation. The drop in reversibility caused by superficial concatenation is a common tendency in our experiments, and the results for all combinations of IT and CS are presented in Appendix~\ref{app:reversibility}.

\subsection{Adding CS and considerate IT selection can be a solution}
\begin{figure}[h]
\centering
\includegraphics[width=\linewidth]{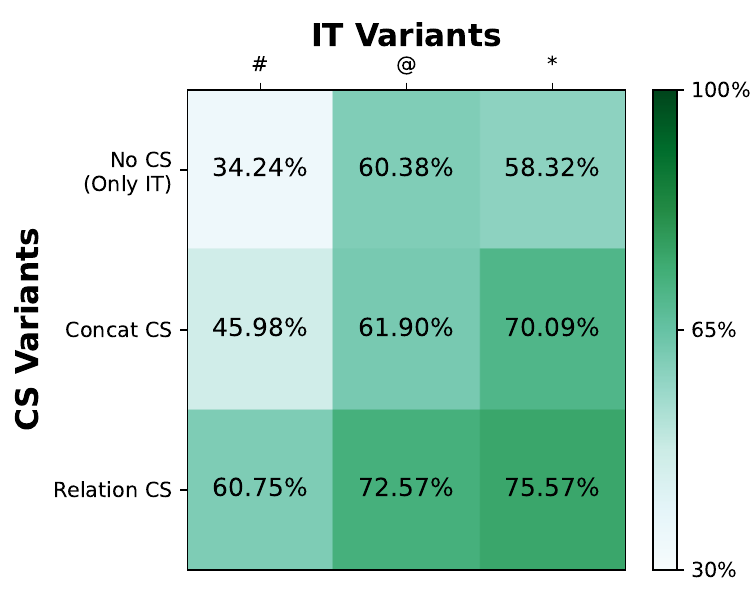}
 \caption{Reversibility after translation for IT and CS variants.} 
 \label{fig:heatmap}
\end{figure}

In this section, we verify that adding CS and prudent selection of IT can relieve the above challenge. We empirically assess the impact of incorporating IT and CS we designed, on the degree of reversibility. Figure~\ref{fig:heatmap} illustrates the average of reversibility over the five-language NLI data translated with NLLB-1.3B, for each case.
As evidenced by our experimental findings, altering IT significantly influences reversibility. Particularly, utilizing `@' as IT can yield over a 25\% increase in reversibility compared to using `\#'. Additionally, the inclusion of CS contributes to enhanced reversibility. Notably, the performance of the Relation CS, which defines clearer relations among components, surpasses that of the Concat CS, which assigns weaker relationships among them. This underscores the effectiveness of our proposed IT+CS methodology in aiding data translation strategies. Further analysis of these impacts is presented in the subsequent sections.

\begin{table*}[h]
\centering
\resizebox{0.92\textwidth}{!}{
\begin{tabular}{c|ccc|c||cc|c}
\toprule[1.5pt]
\multicolumn{1}{c|}{\textbf{Task}} & \multicolumn{4}{c||}{\textbf{WPR}} & \multicolumn{3}{c}{\textbf{QG}}\\ \midrule

\multicolumn{1}{c|}{\textbf{Language}} & \textbf{De} & \textbf{Fr} & \textbf{Zh} & \textbf{Avg} & \textbf{De} & \textbf{Fr} & \textbf{Avg} \\ \midrule[1.5pt]

\multicolumn{1}{c|}{\textbf{Separate}} & 48.630 & 47.491 & 47.620 & 47.913 ( - ) & 24.181 & 25.424 & 24.802 ( - ) \\ \midrule

\textbf{No CS} & 48.420 & 50.146 & 47.462 & 48.676 (+0.763) & 24.733 & 25.715 & 25.224 (+0.422) \\
\textbf{Concat CS} & 48.576 & 50.132 & 47.707 & 48.805 (+0.892) & 24.781 & 25.657 & 25.219 (+0.417) \\
\textbf{Relation CS} & \textbf{50.066} & \textbf{50.593} & \textbf{48.908} & \textbf{49.855 (+1.942)} & \textbf{24.996} & \textbf{25.837} & \textbf{25.416 (+0.614)} \\ \midrule

\multicolumn{8}{c}{\textit{\textbf{Performance of the trained model (vs model trained with individually translated data)}}} \\

\midrule[1.5pt]

\multicolumn{1}{c|}{\textbf{Separate}} & \multicolumn{7}{c}{\textbf{100\%}} \\ \midrule

\textbf{No CS} & 48.91\% & 42.19\% & 31.61\% & 40.90\% (-59.10\%) & 24.35\% & 31.70\% & 28.03\% (-71.97\%) \\
\textbf{Concat CS} & 59.72\% & 61.48\% & 44.81\% & 55.34\% (-44.66\%) & 33.10\% & 31.37\% & 32.24\% (-67.76\%) \\
\textbf{Relation CS} & 82.41\% & 87.61\% & 69.70\% & 79.91\% (-20.09\%) & 35.94\% & 41.82\% & 38.88\% (-61.12\%) \\ \midrule

\multicolumn{8}{c}{\textit{\textbf{Quantity of the training data (vs individually translated data)}}} \\

\bottomrule[1.5pt]

\end{tabular}
}
\caption{Performance of the model trained with each translated dataset. \textbf{Separate} refers to the individual translation of each data component.} \label{tb:res_WPRQG_percent}
\end{table*}

\subsection{IT+CS enhances effectiveness as training data}
We can considerably enhance reversibility through IT+CS, but our MT process inevitably incurs data loss while individual translation of each component would preserve a whole dataset. However, we contend that even though individually translated datasets may exhibit a larger quantity, their quality is likely to be compromised.
Note that the primary focus of this study lies in enhancing the value of translated data as training instances. Considering this, we verify the substantial effectiveness of our approach against the individual translation of each data component, by comparing the performance of the model trained with each translated dataset. 
We report performances of CS variants utilizing `\#' as IT. Experimental results presented in Table~\ref{tb:res_WPRQG_percent} demonstrate the following implications. 

\begin{figure}[t]
\centering
\includegraphics[width=1.02\linewidth]{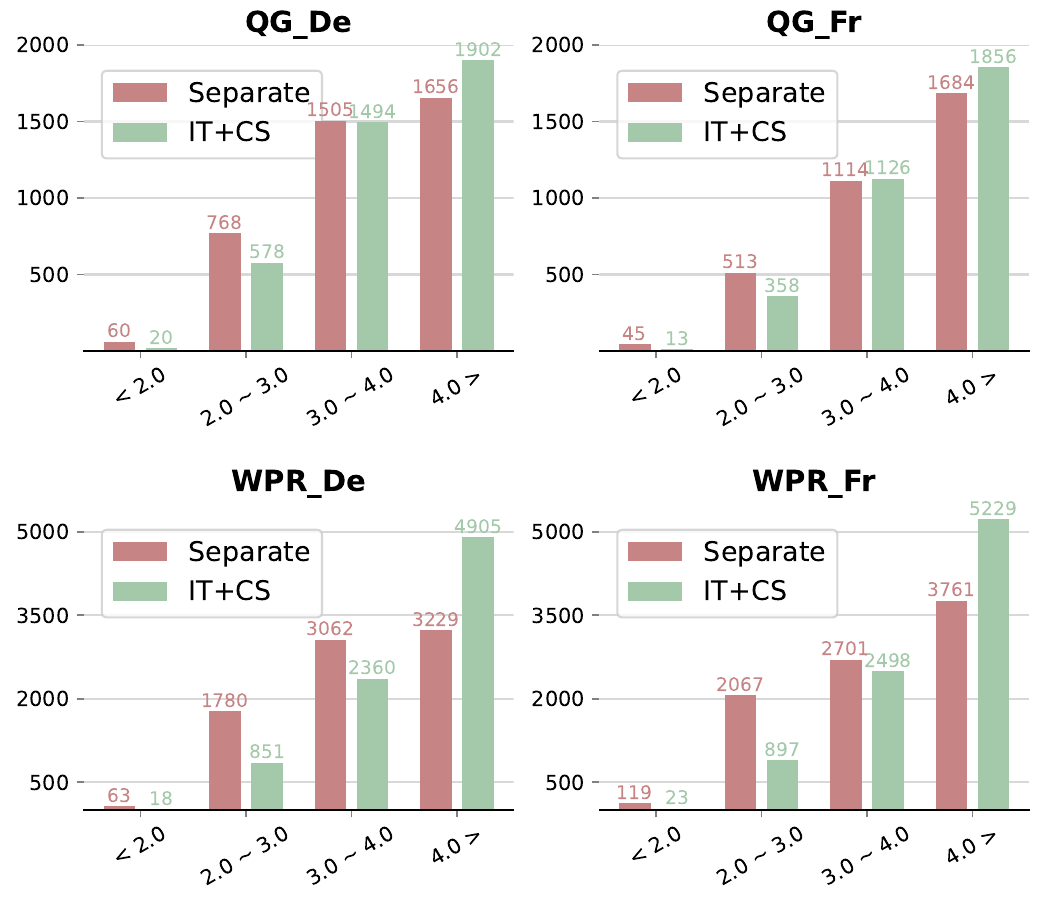}
 \caption{LLM evaluation results. We prompts ChatGPT to provide 0 - 5 scale quality score for each data point. Y-axis represents the quantity of instances which score is in the score range in X-axis.} 
 \label{fig:chatgpt}
\end{figure}

\paragraph{Even small quantity, we can obtain high-quality data}
While fewer in quantity compared to those translated individually for each component, relation-aware translations (\emph{i.e.}, \textbf{No CS}, \textbf{Concat CS} and \textbf{Relation CS}) demonstrated superior performance. Even in cases where only 28\% of the QG data was preserved, the relation-aware translation exhibited greater effectiveness than the 100\% training data generated by translating each component separately. These results validate our framework as an effective pipeline for acquiring high-quality training data.

\paragraph{Relation-aware translation makes better data}
The experimental results demonstrate that all methods concatenating data components for data translation outperform separate translation. Specifically, enhancing the interrelation between data components defined in CS led to improved performance. This underscores the significance of considering inter-component relationships in data translation, as highlighted by our motivation. Particularly we can obtain considerable performance improvement both for QG and WPR, compared to translating each component individually. 

\subsection{LLM Evaluation}
\label{sec:chatgpt}
To elaborate a more meticulous analysis of the impact of the IT+CS strategy on training data translation, we perform a LLM evaluation on the translated data \cite{chen2023alpagasus}. \citet{chen2023alpagasus} proposed an evaluation measure utilizing ChatGPT \cite{OpenAI2022ChatGPT}, that estimates the effectiveness of each data point as a training instance. Drawing inspiration from the previous study, we estimate the utility of each translated dataset as a training source. We adopt GPT3.5-turbo for evaluating each data point. Experimental results are illustrated in Figure~\ref{fig:chatgpt}, with additional details provided in Appendix~\ref{app:chatgpt}.

\begin{figure}[t]
\centering
\includegraphics[width=1.02\linewidth]{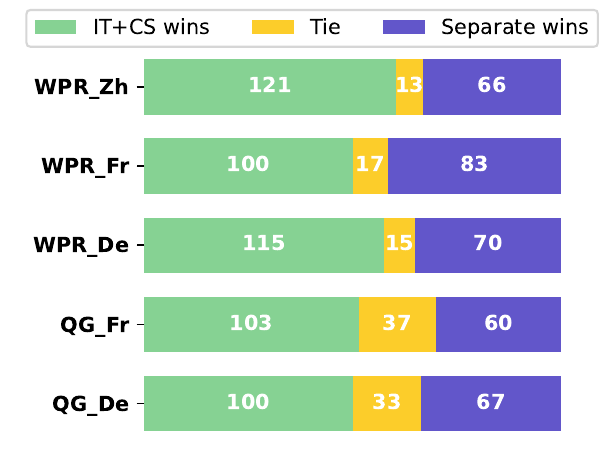}
 \caption{LLM evaluation results. We prompts GPT-4o to provide pairwise evaluation between Separate and IT+CS for each data point.} 
 \label{fig:pairwise}
\end{figure}

As observed from the experimental results, the IT+CS approach significantly increases the proportion of data scoring in the higher range (4.0 >) compared to the method of translating each data individually, while notably reducing the proportion of data scoring in the lower range (2.0 $\sim$ 3.0). This demonstrates the efficacy of our proposed framework in data translation and highlights the vulnerability of strategies translating each data component separately.

Additionally, to facilitate a more intuitive analysis and comparison of the translation results of Separate and IT+CS, we employed GPT-4o for pairwise evaluation. We randomly sampled 200 data points from each dataset and qualitatively compared the translation quality of both approaches. Detailed prompts used for the evaluation are provided in Appendix~\ref{app:chatgpt} and illustrated in Figure~\ref{fig:gpt4}. To minimize the effects of the LLM's positional bias \cite{wang2023large}, we randomized the input order of Separate and IT+CS for each evaluation. The experimental results are presented in Figure~\ref{}. As shown in the figure, the IT+CS strategy yields qualitatively superior translations compared to translating each component separately.

\subsection{IT+CS enhances explicit translation quality}
As XNLI provides human-crafted references for each language, the quality of translated data can explicitly be measured with conventional translation methods. In line with this, we analyze the impact of the relation-considering translation on translation quality. We extract overlapping portions from reversible datasets to form common training data and compare the translation quality and model performance trained on these data. 
\begin{table}[h]
\centering
\resizebox{1.0\linewidth}{!}{
\begin{tabular}{c|ccccc}
\toprule[1.5pt]
\multicolumn{1}{c|}{\textbf{Task}} & \multicolumn{5}{c}{\textbf{NLI}} \\ \midrule

\multicolumn{1}{c|}{\textbf{Language}} & \textbf{De} & \textbf{Zh} & \textbf{Fr} & \textbf{Hi} & \textbf{Vi} \\ \midrule[1.5pt]

\textbf{Separate} & 38.887 & 26.634 & 56.281 & 14.004 & 41.796 \\ \midrule
\textbf{No CS} & 39.456 & 26.455 & 56.664 & 14.147 & 42.437 \\
\textbf{Concat CS} & 38.398 & 26.474 & 56.598 & \textbf{14.437} & 42.966 \\
\textbf{Relation CS} & \textbf{39.774} & \textbf{26.714} & \textbf{57.006} & 14.425 & \textbf{43.094} \\ \midrule

\multicolumn{6}{c}{\makecell[c]{
\textit{\textbf{BLEU score of the common training data}}
}}\\ 

\midrule[1.5pt]

\textbf{Separate} & 66.607 & \textbf{64.830} & 68.862 & \textbf{67.605} & 66.846 \\ \midrule
\textbf{No CS} & 65.589 & 64.088 & \textbf{69.840} & 65.329 & 69.062 \\
\textbf{Concat CS} & 67.505 & 62.495 & 69.561 & 65.449 & 69.741 \\
\textbf{Relation CS} & \textbf{68.204} & 64.092 & 68.543 & 65.230 & \textbf{70.220} \\ \midrule
\multicolumn{6}{c}{\makecell[c]{
\textit{\textbf{Performance of the trained model}}  
}}\\ 
\bottomrule[1.5pt]

\end{tabular}
}
\caption{Performance of the model trained with each translated dataset. We derive a common index set for each language that intersects all the translated datasets. Then we extract a subset for each translated dataset, which indices are all included in the common index set.} \label{tb:res_NLI_bleu_nllb}
\end{table}

As shown in Table~\ref{tb:res_NLI_bleu_nllb}, IT+CS significantly improves translation quality. This improvement is particularly pronounced in alphabetic languages, with Vietnamese showing a 1.298-point enhancement in translation quality and a corresponding 3.374-point improvement in model performance compared to the Separate translation approach. However, performance degradation is observed in non-alphabetic languages, likely attributable to MT performance itself. We plan further analysis on this phenomenon. 

\subsection{IT+CS on the MT model variants}
To verify the general applicability of the framework we propose, we evaluate the performance across three different models. The experimental results are presented in Table~\ref{tb:model_variants}. 
\begin{table}[h]
\centering
\resizebox{1.0\linewidth}{!}{
\begin{tabular}{l||cccccc}
\toprule[1.5pt]

\multicolumn{1}{c}{} & \multicolumn{2}{c}{\textbf{NLLB-600M}} & \multicolumn{2}{c}{\textbf{NLLB-1.3B}} & \multicolumn{2}{c}{\textbf{M2M-418M}} \\
\multicolumn{1}{c}{} & \textbf{Separate} & \textbf{IT+CS} & \textbf{Separate} & \textbf{IT+CS} & \textbf{Separate} & \textbf{IT+CS} \\ \midrule[1.5pt] 

\multicolumn{7}{c}{\textbf{WPR}} \\ \midrule

\textbf{De} & 48.630 & \textbf{49.519} & 47.560 & \textbf{50.710} & 48.950 & \textbf{49.845} \\
\textbf{Fr} & 47.491 & \textbf{48.729} & 46.721 & \textbf{49.208} & 49.060 & \textbf{50.151} \\
\textbf{Zh} & 47.620 & \textbf{49.056} & 46.891 & \textbf{49.326} & 47.301 & \textbf{49.191} \\
\textbf{Avg} & 47.913 & \textbf{49.101} & 47.057 & \textbf{49.748} & 48.437 & \textbf{49.729} \\ \midrule [1.5pt]

\multicolumn{7}{c}{\textbf{QG}} \\ \midrule

\textbf{De} & 24.181 & \textbf{25.420} & 25.535 & \textbf{25.607} & 23.960 & \textbf{25.135} \\
\textbf{Fr} & 25.424 & \textbf{25.876} & 25.833 & \textbf{26.053} & 24.676 & \textbf{25.970} \\
\textbf{Avg} & 24.802 & \textbf{25.648} & 25.684 & \textbf{25.830} & 24.318 & \textbf{25.552} \\

\bottomrule[1.5pt]

\end{tabular}
}
\caption{Experiment on model variants. We report the performance of the model trained with each translated dataset.} \label{tb:model_variants}
\end{table}

Here, we report on the performance of `*' IT, which consistently exhibits high performance across all three models. Specifically, we can obtain 2.690 and 0.845 point performance improvement, for each WPR and QG. As can be seen from our experimental results, our method outperforms the separate translation approach in terms of performance across all MT systems, all datasets, and languages. This validates the broad applicability of our method as a data translation framework.

\subsection{Qualitative Analysis}
To delve deeper into the effectiveness of IT+CS in data translation and for the further analyses of translation results, we examine actual translation outcomes along with cross attention map analysis. The results are described in Appendix~\ref{app:qualitive}. Through our experimental results, we affirm the practical superiority of the IT+CS strategy in its application to data translation, especially for the multifaceted data structure.

\section{Conclusion}
This study explored challenges encountered when implementing data translation through MT frameworks. We highlighted that individual translation of each data component neglects their interrelations, leading to a compromise in data quality. While composing a singular sequence by concatenating all the components theoretically can alleviate this, it also introduces the limitation of the inability to restore data components from the translated sequence. As a solution, we introduced a relation-considering translation pipeline that integrates IT and CS. This approach led to a substantial enhancement in the quality of training data as opposed to separate component translation. Our empirical findings underscored the paramount importance of inter-component relation in data translation, emphasizing that considering this relation can facilitate high-level data translations. This progression lays a foundation for future data translation research.

\section*{Limitation}
We identify three potential constraints of our experimental setting.
Firstly, variants of IT and RP were only tested under three specific cases. We were unable to validate every possible case, and there may exist other optimal types of IT or RP. While it is challenging to claim our results as optimal, our experiment conclusively affirmed that even subtle changes in IT can lead to evident performance improvement, and reinforcing the interrelation within each data component by concatenating RP can result in superior quality training data. Our experimental design encapsulates sufficient discussion to reach this conclusion.

The second limitation pertains to the variants of NMT models. We employed only three types of NMT models. Testing against a wider array of translation models could significantly enhance the general applicability of our study, but this was hindered by our resource constraints. Nevertheless, our experiments cover the difference in the model size (NLLB-600M and NLLB-1.3B) and the difference in the NMT training data or training strategy (NLLB and M2M) to induce more generalizable results.

Lastly, we confined language variants in our experiments. Due to resource constraints, we could not experiment with all languages provided by XGLUE and XNLI. However, we set up more than two languages for each task, to ensure our results not be biased towards any specific language. We deemed the varied performance and tendencies across different languages within NLI as a significant discovery. We did not perform further analyses as such discovery may fall beyond the scope of this paper, but we present an interesting scope for future research.


\section*{Ethics Statement}
We utilized the publicly available XGLUE benchmark and XNLI datasets. We adhere strictly to the copyright of the original research in relation to the language resources and translated data used. Given that the utility and validity of XLGUE and XNLI have been established in numerous prior studies, we confirm that there were no distinct ethical issues encountered in our usage of these datasets.


\section*{Acknowledgements}

This work was partly supported by ICT Creative Consilience Program through the Institute of Information \& Communications Technology Planning \& Evaluation(IITP) grant funded by the Korea government(MSIT) (IITP-2024-RS-2020-II201819, 20\%), Institute for Information \& communications Technology Promotion(IITP) grant funded by the Korea government(MSIT) (RS-2024-00398115, Research on the reliability and coherence of outcomes produced by Generative AI, 40\%) and Basic Science Research Program through the National Research Foundation of Korea(NRF) funded by the Ministry of Education(NRF-2021R1A6A1A03045425, 40\%).

\bibliography{custom}

\appendix

\section{Implementation Details}
All experiments were performed using the RTX A6000. A single GPU was utilized for the training of an individual model, with early stopping criteria applied within 20 epochs. The learning rates selected for all tests were chosen among 1e-04, 3e-05, 5e-05, or 1e-04. The XLM-r base model was fine-tuned using a lr of 2e-05, while the mT5-base model employed a lr of 1e-05. HuggingFace \cite{wolf2020transformers} provided the foundation for all model configurations and training pipeline.


\label{app:implementation}
\section{Dataset Details}
We validate the effectiveness of our approach with the XNLI dataset \cite{conneau2018xnli} and selected two tasks in the XGLUE benchmark \cite{liang2020xglue}. The \textbf{NLI} dataset comprises pairs of sentences with a label categorizing the semantic relationship between the two sentences into one of three classifications: entailment, contradiction, or neutral. This task aims to illustrate the effectiveness of considering semantic relationships during translations. 

In the \textbf{WPR} task, the goal is to predict the relevance of a web page to a given query. Each instance is a 4-part tuple: query, web page title, web page snippet, and label. The relevance label includes ratings from Perfect (4) to Bad (0). This task is included to verify the effectiveness of our approach in dealing with more than two components. 

\textbf{QG} is a generation task, comprising a passage and a question that could originate from the given passage.  In this case, we investigate the generalizability of our approach for lengthier translation units. To acquire more general results, we conduct experiments in five languages: English (En), Chinese (Zh), French (Fr), Vietnamese (Vi), and Hindi (Hi). These were chosen based on generally conceived resource quantity differences and the shared alphabetic character system. We validate all five languages for NLI. We select three existing languages~(De, Fr, Zh) among the five for WPR, and two languages~(De, Fr) for QG. Detailed statistics of each dataset are described in Table~\ref{tb:data_stat}.

As we take the simplest form of IT, our experimented IT types can be included in the training dataset. To address this, we elaborate the count of data points that include each IT in Table~\ref{tb:data_stat}. In comparison to the total data volume, the counts of data containing IT are deemed negligible. Given the data reversibility is not 100\%, we postulate that the bias resulting from omitted data will likely be minimal.

\begin{table}[h]
\centering
\resizebox{0.42\textwidth}{!}{
\begin{tabular}{lccc}
\toprule[1.5pt]
\makecell[c]{\textbf{Task}} & \textbf{XNLI} & \textbf{WPR} & \textbf{QG} \\ \midrule[1.5pt]
\multicolumn{4}{c}{\textbf{Train}} \\  \midrule
\textbf{Num of data} & 392,702 & 99,997 & 100,000 \\
\textbf{Containing} \text{\#} & 55 & 2,101 & 709 \\
\textbf{Containing} \text{@} & 80 & 1,558 & 79 \\
\textbf{Containing} \text{*} & 66 & 998 & 374 \\ \midrule
\multicolumn{4}{c}{\textbf{Validation}} \\ \midrule
\textbf{Num of data} & 2,490 & 10,008 & 10,000 \\
\textbf{Containing} \text{\#} & 0 & 240 & 25 \\
\textbf{Containing} \text{@} & 0 & 144 & 13 \\
\textbf{Containing} \text{*} & 0 & 109 & 34 \\ \midrule
\multicolumn{4}{c}{\textbf{Test}} \\ \midrule
\textbf{Num of data} & 5,010 & 10,004 & 10,000 \\
\textbf{Containing} \text{\#} & 3 & 234 & 30 \\
\textbf{Containing} \text{@} & 0 & 167 & 11 \\
\textbf{Containing} \text{*} & 0 & 107 & 46 \\
\bottomrule[1.5pt]

\end{tabular}
}
\caption{Data statistics.} \label{tb:data_stat}
\end{table}

\label{app:datastat}
\section{LLM Evaluation Details}

\begin{table*}[h]
\centering
\resizebox{0.95\textwidth}{!}{
\begin{tabular}{cc|ccccc|cc|ccc|c}
\toprule[1.5pt]
\multicolumn{2}{c|}{\textbf{Task}} & \multicolumn{5}{c|}{\textbf{NLI}} & \multicolumn{2}{c|}{\textbf{QG}} & \multicolumn{3}{c|}{\textbf{WPR}} & \multirow{2}{*}{\textbf{Avg}}\\ \cmidrule{1-12}

\multicolumn{2}{c|}{\textbf{Language}} & \textbf{De} & \textbf{Zh} & \textbf{Fr} & \textbf{Hi} & \textbf{Vi} & \textbf{De} & \textbf{Fr} & \textbf{De} & \textbf{Fr} & \textbf{Zh} & \\ \midrule[1.5pt]
 
\multirow{3}{*}{\textbf{@}} & \textbf{No CS} & 70.36\% & 32.63\% & \textbf{65.89\%} & 80.63\% & 73.49\% & 39.88\% & \textbf{43.82\%} & 85.20\% & 72.00\% & 39.96\% & 60.39\% \\
 & \textbf{Concat CS} & 81.04\% & \textbf{56.18\%} & 48.27\% & \textbf{96.52\%} & 63.76\% & 43.33\% & 30.73\% & 79.43\% & 69.16\% & 50.72\% & 61.91\% \\
 & \textbf{Relation CS} & \textbf{85.63\%} & 52.38\% & 65.33\% & 95.74\% & \textbf{74.20\%} & \textbf{61.16\%} & 42.29\% & \textbf{87.68\%} & \textbf{88.51\%} & \textbf{73.00\%} & \textbf{72.59\%} \\ \midrule
 
\multirow{3}{*}{\textbf{\#}} & \textbf{No CS} & 14.86\% & 11.64\% & 18.23\% & 64.71\% & 54.33\% & 24.35\% & 31.70\% & 48.91\% & 42.19\% & 31.61\% & 34.25\% \\
 & \textbf{Concat CS} & 25.81\% & 25.62\% & 27.72\% & 79.34\% & \textbf{70.94\%} & 33.10\% & 31.37\% & 59.72\% & 61.48\% & 44.81\% & 45.99\% \\
 & \textbf{Relation CS} & \textbf{46.44\%} & \textbf{36.12\%} & \textbf{54.07\%} & \textbf{82.69\%} & 70.86\% & \textbf{35.94\%} & \textbf{41.82\%} & \textbf{82.41\%} & \textbf{87.61\%} & \textbf{69.70\%} & \textbf{60.77\%} \\ \midrule
 
\multirow{3}{*}{\textbf{*}} & \textbf{No CS} & 40.27\% & 25.09\% & 41.90\% & 83.60\% & 71.20\% & 51.39\% & 53.18\% & 79.33\% & 76.03\% & 61.35\% & 58.33\% \\
 & \textbf{Concat CS} & 68.63\% & 51.70\% & 57.50\% & \textbf{91.12\%} & \textbf{78.08\%} & 61.65\% & 58.42\% & 80.73\% & 80.59\% & 72.61\% & 70.10\% \\
 & \textbf{Relation CS} & \textbf{75.08\%} & \textbf{55.52\%} & \textbf{66.78\%} & 89.16\% & 75.94\% & \textbf{69.01\%} & \textbf{70.80\%} & \textbf{84.56\%} & \textbf{90.83\%} & \textbf{78.13\%} & \textbf{75.58\%} \\

\bottomrule[1.5pt]

\end{tabular}
}
\caption{Percentage of data reversibility after translation under NLLB-600M.} \label{tb:res_percent}
\end{table*}

In our study, we leverage LLMs to assess the quality of datasets translated through various methodologies. \citet{zheng2023judging} indicate that LLMs' ability to align with preferences identified through both controlled experiments and crowdsourced methodologies exhibits a remarkable concordance rate exceeding 80\%. This evidence underscores the potential of advanced language models in reflecting human judgments. Furthermore, following the methodology described by \citet{chen2023alpagasus} in utilizing LLMs as ChatGPT for data quality assessment, applying filtering criteria based on LLM evaluations resulted in a significant reduction of low-quality datasets, leading to improved performance of the trained model. This outcome serves as evidence of the effectiveness of employing LLMs for data quality assessment purposes.

To tailor the evaluation process to our specific needs, we adapted the prompts from \cite{chen2023alpagasus} to assess the quality of our translated datasets and employed GPT-3.5-turbo as our evaluator. We conducted quality assessments on sentences translated from English source sentences in the XGLUE dataset's test set to De, Zh, and Fr using our method. The prompts used for each task were customized to reflect our evaluation criteria, illustrating the adaptability and precision of our methodology in assessing translation quality across diverse data contexts.

For pairwise comparison, we use the prompt shown in Figure~\ref{fig:gpt4}. To minimize the effects of LLM's positional bias \cite{wang2023large}, we randomly set the input order of Separate and IT+CS for each evaluation. Additionally, we designed the evaluation setup with a more refined prompt to ensure a more objective assessment.

\definecolor{yello}{HTML}{FDF3D0}
\definecolor{boldyellow}{HTML}{B89230}

\definecolor{green}{HTML}{E5F0DB}
\definecolor{boldgreen}{HTML}{6F8F52}

\definecolor{blue}{HTML}{D2E0FB}
\definecolor{boldblue}{HTML}{8EACCD}

\begin{figure}[!h]
\begin{mdframed}[linewidth=1.5pt,roundcorner=14pt,backgroundcolor=yello,middlelinecolor=boldyellow]
\small 
\textbf{System Prompt:}
\begin{verbatim}
We would like to request your feedback on the 
performance of AI assistant in response to the
passage and the given question displayed 
following.
                
passage: [passage]
question: [question]
    \end{verbatim}

\textbf{User Prompt:}
\begin{verbatim}
Please rate according to the [dimension] of the 
response to the passage and the question. Each 
assistant receives a score on a scale of 0 to 5, 
where a higher score indicates a higher level of 
the [dimension]. 
Please first output a single line containing 
the value indicating the scores. In the 
subsequent line, please provide a comprehensive 
explanation of your evaluation, avoiding any 
potential bias.
\end{verbatim}
\end{mdframed}
\caption{Prompt template for ChatGPT evaluation of QG task. We evaluated each data point with the 0-5 scale quality score.}
\end{figure}

\begin{figure}[!h]
\begin{mdframed}[linewidth=1.7pt,roundcorner=14pt,backgroundcolor=green,middlelinecolor=boldgreen]
\small 
\textbf{System Prompt:}
\begin{verbatim}
We would like to request your feedback on the 
performance of AI assistant in response to 
the query and the given title and snippet 
displayed following
                
query: [query]
title: [title]
snippet: [snippet]
    \end{verbatim}

\textbf{User Prompt:}
\begin{verbatim}
Please rate according to the {dimension} of the
response to the passage and the question. Each 
assistant receives a score on a scale of 0 to 5,
where a higher score indicates higher level of 
the {dimension}. 
Please first output a single line containing 
the value indicating the scores. In the 
subsequent line, please provide a comprehensive
explanation of your evaluation, avoiding any 
potential bias.
\end{verbatim}
\end{mdframed}
\caption{Prompt template for ChatGPT evaluation of WPR task. We evaluated each data point with the 0-5 scale quality score.}
\end{figure}

\begin{figure*}[!h]
\begin{mdframed}[linewidth=1.5pt,roundcorner=14pt,backgroundcolor=blue,middlelinecolor=boldblue]
\small 
\textbf{System Prompt:}
\begin{verbatim}
Please act as an impartial judge and evaluate the quality of the two statements.
You will be given a English source statement and two {language} translated statements. 
Each statement is generated by an AI translation assistant.
You should choose the statement that correctly evaluated each response and provided better quality 
explanation to their assessment.
Your evaluation should consider factors such as the correctness, fluency, relevance, 
accuracy and coherence.
Begin your evaluation by comparing the two statements and provide a short explanation.
Avoid any position biases and ensure that the order in which the responses were presented does not 
influence your decision. 
Do not allow the length of the responses to influence your evaluation. 
Do not favor certain names of the assistants. 
Be as objective as possible. 
After providing your explanation, output your final verdict by strictly following this format: 
"[[A]]" if Statement A is better, "[[B]]" if Statement B is better, and "[[C]]" for a tie.""",

    \end{verbatim}

\textbf{User Prompt:}
\begin{verbatim}
[Source]
{source}
[The Start of Statement A]
{translation 1}
[The End of Statement A]
[The Start of Statement B]
{translation 2}
[The End of Statement B]
\end{verbatim}
\end{mdframed}
\caption{Prompt template for pairwise comparison by GPT4o. We compare translation results derived from \textbf{Separate} and \textbf{IT+CS} strategies. \texttt{\{source\}} denotes source statement and \texttt{\{translation 1\}} and \texttt{\{translation 2\}} denote its transltaion resutls. \texttt{\{source\}} is set to be English, and each translation is in \texttt{\{language\}} language.
} \label{fig:gpt4}
\end{figure*}
\label{app:chatgpt}

\section{Data Reversibility}

When employing a translation model to translate data, reversibility is a crucial factor. High reversibility directly impacts the number of translated data instances and contributes to increasing the variability of data during model training. We experimented with all combinations of relation prompts and indicator tokens.

As indicated in Table~\ref{tb:res_percent}, regardless of the type of indicator token used, leveraging relation prompt results in significantly high average reversibility. Notably, when the `\#' indicator token was used, reversibility improved by approximately 77\% when considering relations. However, the No RP scenario shows the lowest level of preservation during the translation process. This trend is also observed in other translation models we adopted. This suggests that leveraging a relation-considered relation prompt in translation can be an appropriate means to secure more data amount, regardless of the language or task.



\label{app:reversibility}

\section{Qualitive Analysis}
\begin{table*}[ht]
\centering
\small
\begin{tabular}{c|l}
\toprule[1.5pt]
\multicolumn{2}{c}{\textbf{Sample 1 - QG}} \\ \midrule

\makecell[c]{\textbf{Before} \\ \textbf{Translation}} & 
\makecell[l]{
- \textbf{passage}: Markice Moore as Andrew on The Walking Dead (AMC) Markice Moore, an actor best known \\ for his role as Andrew on season three of The Walking Dead, is recovering at home after being \\ hospitalized for a suicide attempt, according\\ to The Hollywood Reporter. \\
- \textbf{question}: walking dead star suicide attempt
} \\ \midrule

\makecell[c]{\textbf{Separate}} & 
\makecell[l]{
- \textbf{passage}: Markice Moore als Andrew in The Walking Dead (AMC) Markice Moore, ein Schauspieler, \\der für seine Rolle als Andrew in der dritten Staffel von The Walking Dead bekannt ist, erholt sich nach \\einem Selbstmordversuch nach dem Krankenhaus nach dem Hollywood Reporter. \\
- \textbf{question}: Selbstmordversuch mit einem starken Sternen
} \\ \midrule

\makecell[c]{\textbf{IT only}} & 
\makecell[l]{
Markice Moore als Andrew in The Walking Dead (AMC) Markice Moore, ein Schauspieler, der am besten \\ für seine Rolle als Andrew in der dritten Staffel von The Walking Dead bekannt ist, erholt sich \\nach einem Selbstmordversuch im Krankenhaus.
} \\ \midrule

\makecell[c]{\textbf{IT+CS}} & 
\makecell[l]{
- \textbf{passage}: Markice Moore als Andrew in The Walking Dead (AMC) Markice Moore, ein Schauspieler,\\der am besten für seine Rolle als Andrew in der dritten Staffel von The Walking Dead bekannt ist, erholt \\sich nach einem Selbstmordversuch nach dem Krankenhaus nach dem Hollywood Reporter. \\
- \textbf{question}: Walking Dead Star Selbstmordversuch
} \\ \midrule[1.5pt]


\multicolumn{2}{c}{\textbf{Sample 2 - WPR}} \\ \midrule

\makecell[c]{\textbf{Before} \\ \textbf{Translation}} & 
\makecell[l]{
- \textbf{query}: twitw \\
- \textbf{title}: Twitter Developer Platform — Twitter Developers \\
- \textbf{snippet}: Twitter is the best place in the world for businesses and people to connect. Since the \\early days of Twitter people have used the public, live, and conversational nature of the platform to \\engage with businesses.
} \\ \midrule

\makecell[c]{\textbf{Separate}} & 
\makecell[l]{
- \textbf{query}: Schlagwort \\
- \textbf{title}: Twitter-Entwickler-Plattform  Twitter-Entwickler \\
- \textbf{snippet}: Seit den frühen Tagen von Twitter nutzen Menschen die öffentliche, live und konversationsartige \\Natur der Plattform, um mit Unternehmen zu interagieren.
} \\ \midrule

\makecell[c]{\textbf{IT only}} & 
\makecell[l]{
- Twitter ist der beste Ort der Welt für Unternehmen und Menschen, um sich zu verbinden. Seit den frühen \\Tagen von Twitter haben die Menschen die öffentliche, live und konversative Natur der Plattform genutzt, \\um mit Unternehmen zu interagieren.
} \\ \midrule

\makecell[c]{\textbf{IT+CS}} & 
\makecell[l]{
- \textbf{query}: twitw \\
- \textbf{title}: Twitter Developer Platform  Twitter Developers \\
- \textbf{snippet}: Twitter ist der beste Ort der Welt für Unternehmen und Menschen, um sich zu verbinden. \\Seit den frühen Tagen von Twitter haben die Menschen die öffentliche, \\live und konversative Natur der Plattform verwendet, um mit Unternehmen zu interagieren.
} \\

\bottomrule[1.5pt]

\end{tabular}
\caption{System-level qualitative analysis.}
\label{tab:system}
\end{table*}

\begin{table*}[ht]
\centering
\renewcommand\arraystretch{2} 
\setlength\extrarowheight{-2pt} 
\small
\begin{tabular}{c|l|l}
\toprule[1.5pt]
\multicolumn{1}{c}{} & \multicolumn{2}{c}{\textbf{Reference(En)}}\\ \cmidrule(lr){2-3}

\multirow{1}{*}{\textbf{Query}} & \multicolumn{2}{l}{100\% cotton racerback camisoles 1XL}\\\cmidrule(lr){2-3}
\multirow{1}{*}{\textbf{Title}} & \multicolumn{2}{l}{Sofra Women's 100\% Cotton Racerback Tank Top - amazon.com}\\\cmidrule(lr){2-3}
\multirow{2}{*}{\textbf{Snippet}}  & \multicolumn{2}{l}{\parbox[t]{13cm}{Sofra Women's 100\% Cotton Racerback Tank Top ... A bit worried what will happen when I wash them as they are 100\% cotton. I think they will be ok to wear under other things, the fabric is slightly shear so probably not good with nothing over it. You get what you pay for is the lesson here. Read more. Helpful.}}\\

\cmidrule(lr){2-3}

 \multicolumn{1}{c}{} & \multicolumn{1}{c}{\textbf{Separate(De)}} & \multicolumn{1}{c}
 {\textbf{IT+CS(De)}} \\
 
\cmidrule(lr){2-2} \cmidrule(lr){3-3}

\multirow{1}{*}{\textbf{Query}}  & 100\% Baumwoll-Rennstreifenhemden 1XL & 100\% Baumwoll-Racerback-Shemisoles 1XL\\\cmidrule(lr){2-2} \cmidrule(lr){3-3}

\multirow{1}{*}{\textbf{Title}} & \parbox[t]{6.5cm}{Sofra Frauen 100\% Baumwoll-Racerback Tank Top - amazon.com }& \parbox[t]{6.5cm}{Sofra Frauen 100\% Baumwoll-Racerback Tank Top - amazon.com }\\\cmidrule(lr){2-2} \cmidrule(lr){3-3}

\multirow{5}{*}{\textbf{Snippet}}
& \parbox[t]{6.5cm}{ Sofra Frauen 100\% Cotton Racerback Tank Top... ein wenig besorgt, was passiert, wenn ich sie wasche, da sie 100\% Baumwolle sind. Ich denke, sie werden in Ordnung sein, um unter andere Dinge zu tragen, der Stoff ist leicht scheren, so wahrscheinlich nicht gut mit nichts über. Sie bekommen, was Sie bezahlen ist die Lektion hier. Lesen Sie mehr. hilfreich. } 

& \parbox[t]{6.5cm}{ Sofra Women's 100\% Cotton Racerback Tank Top... Ein bisschen besorgt, was passiert, wenn ich sie wasche, da sie 100\% Baumwoll sind. Ich denke, sie werden in Ordnung sein, um unter anderen Dingen zu tragen, der Stoff ist leicht scheren, so dass wahrscheinlich nicht gut mit nichts darüber. Sie bekommen, was Sie bezahlen ist die Lektion hier. Lesen Sie mehr. Hilfreich. } \\
\cmidrule(lr){2-2}\cmidrule(lr){3-3}
\multirow{1}{*}{\textbf{LLM Score}} & \makecell[c]{\textbf{2}} & \makecell[c]{\textbf{4}} \\
\midrule[1.5pt]

\multicolumn{1}{c}{} & \multicolumn{2}{c}{\textbf{Reference(En)}}\\\cmidrule(lr){2-3}

\multirow{1}{*}{\textbf{Query}} & \multicolumn{2}{l}{ llc online application for florida }\\ \cmidrule(lr){2-3}
\multirow{1}{*}{\textbf{Title}} & \multicolumn{2}{l}{ Corporations - Division of Corporations - Florida ...  }\\\cmidrule(lr){2-3}
\multirow{2}{*}{\textbf{Snippet}}   & \multicolumn{2}{l}{\parbox[t]{13cm}{ Make all checks payable to the Florida Department of State. Check and money orders must be payable in U.S. currency drawn from a U.S. bank. Credit cards accepted for filing online are MasterCard, Visa, Discover and American Express. Prepaid Sunbiz E-File Account. Processing. File online: 2-3 business days. }}\\

\cmidrule(lr){2-3}

 \multicolumn{1}{c}{} & \multicolumn{1}{c}{\textbf{Separate(De)}} & \multicolumn{1}{c}
 {\textbf{IT+CS(De)}} \\
 
 \cmidrule(lr){2-2} \cmidrule(lr){3-3}
 
\multirow{1}{*}{\textbf{Query}} & llc Online-Bewerbung für Florida & llc Online-Anwendung für Florida \\\cmidrule(lr){2-2} \cmidrule(lr){3-3}

\multirow{1}{*}{\textbf{Title}} & \parbox[t]{6.5cm}{ Unternehmen - Unternehmensbereich - Florida... }& \parbox[t]{6.5cm}{ Corporations - Division of Corporations - Florida... }\\\cmidrule(lr){2-2} \cmidrule(lr){3-3}

\multirow{5}{*}{\textbf{Snippet}}
& \parbox[t]{6.5cm}{ Alle Schecks müssen an das Florida-Außenministerium gezahlt werden. Schecks und Geldbestellungen müssen in US-Währung aus einer US-Bank gezahlt werden. Kreditkarten, die für die Online-Aufgabe akzeptiert werden, sind MasterCard, Visa, Discover und American Express. Prepaid Sunbiz E-File Account. Verarbeitung. Online-Aufgabe: 2-3 Werktage. } 

& \parbox[t]{6.5cm}{ Alle Schecks an das Florida State Department zu zahlen machen. Schecks und Geldbefehle müssen in US-Währung aus einer US-Bank gezogen werden. Kreditkarten, die für die Online-Aufgabe akzeptiert werden, sind MasterCard, Visa, Discover und American Express. Prepaid Sunbiz E-File Account. Verarbeitung. Online-Datei: 2-3 Werktage. } \\
\cmidrule(lr){2-2}\cmidrule(lr){3-3}
\multirow{1}{*}{\textbf{LLM Score}} & \makecell[c]{\textbf{2}} & \makecell[c]{\textbf{4}} \\

\bottomrule[1.5pt]
\end{tabular}
\caption{Samples of translations result using the ``Separate'' and ``IT+RP'' method for the WPR dataset.}
\label{tab:wpr}
\end{table*}

Table~\ref{tab:system}, Table~\ref{tab:wpr} and Table~\ref{tab:qg} show the sample of translated components of each dataset using NLLB-600M. It compares the ``Separate'' and ``IT+CS'' methods, with CS using `\#' symbol. It indicates that the original English text was translated into German. Additionally, LLM Eval Score represents the results obtained using the approach detailed in Section \ref{sec:chatgpt}. Table~\ref{tab:wpr} presents exemplars of results for each translation method applied to the WPR dataset. In terms of contextual similarity and terminological consistency, ``IT+CS'' outperforms ``Separate'' offering a more faithful interpretation of the original text.

\begin{table*}[ht]
\centering
\renewcommand\arraystretch{2} 
\setlength\extrarowheight{-2pt} 
\small
\begin{tabular}{c|l|l}
\toprule[1.5pt]

\multicolumn{1}{c}{} & \multicolumn{2}{c}{\textbf{Reference(En)}}\\ \cmidrule(lr){2-3}

\multirow{1}{*}{\textbf{Passage}}  & \multicolumn{2}{l}{\parbox[t]{13cm}{ There is Frost on the Coils. The coils you see on the back of your dehumidifier are called the evaporator. When the unit is running, these coils turn ice cold. As the fan draws the moist air over the coils, the humidity in the air condenses on these coils. }}\\\cmidrule(lr){2-3}
\multirow{1}{*}{\textbf{Question}} & \multicolumn{2}{l}{why do dehumidifiers stop working}\\\cmidrule(lr){2-3}

 \multicolumn{1}{c}{} & \multicolumn{1}{c}{\textbf{Separate(De)}} & \multicolumn{1}{c}
 {\textbf{IT+CS(De)}} \\\cmidrule(lr){2-2}\cmidrule(lr){3-3}

\multirow{1}{*}{\textbf{Passage}}
& \parbox[t]{6.5cm}{ Die Spulen, die Sie auf der Rückseite Ihres Entfeuchters sehen, werden als Verdampfer bezeichnet. Wenn die Einheit läuft, werden diese Spulen zum Eis kalt. } 

& \parbox[t]{6.5cm}{ Es gibt Frost auf den Coils. Die Coils, die Sie auf der Rückseite Ihres Entfeuchters sehen, werden als Verdampfer bezeichnet. Wenn die Einheit läuft, werden diese Coils zum Eis kalt. Wenn der Lüfter die feuchte Luft über die Coils zieht, kondensiert sich die Luftfeuchtigkeit auf diesen Coils. } \\\cmidrule(lr){2-2}\cmidrule(lr){3-3}

\multirow{1}{*}{\textbf{Question}} & \parbox[t]{6cm}{Warum die Entfeuchter nicht mehr funktionieren} & 
{Warum hören Entfeuchter auf zu arbeiten}\\\cmidrule(lr){2-2}\cmidrule(lr){3-3}
\multirow{1}{*}{\textbf{LLM Score}} & \makecell[c]{\textbf{2}} & \makecell[c]{\textbf{4}} \\

\midrule[1.5pt]

\multicolumn{1}{c}{} & \multicolumn{2}{c}{\textbf{Reference(En)}}\\\cmidrule(lr){2-3}

\multirow{1}{*}{\textbf{Passage}}  & \multicolumn{2}{l}{\parbox[t]{13cm}{ falls on a Saturday and as such, the due date for New Hampshire Interest \& ... and Business Tax returns will be due on Tuesday April 18, 2017. Return due dates for all other tax types with a due date of April 15th are not impacted by ... }}\\\cmidrule(lr){2-3}
\multirow{1}{*}{\textbf{Question}} & \multicolumn{2}{l}{  nh business tax due date  }\\\cmidrule(lr){2-3}

 \multicolumn{1}{c}{} & \multicolumn{1}{c}{\textbf{Separate(De)}} & \multicolumn{1}{c}
 {\textbf{IT+CS(De)}} \\\cmidrule(lr){2-2} \cmidrule(lr){3-3}

\multirow{1}{*}{\textbf{Passage}}   
& \parbox[t]{6.5cm}{ Die Frist für die Erstattung von Zinsen und... und Unternehmenssteuer wird am Dienstag, 18. April 2017 verfallen. Die Frist für die Erstattung aller anderen Steuertypen mit Ablaufdatum vom 15. April wird nicht von... } 

& \parbox[t]{6.5cm}{fällt am Samstag und als solches wird das Fälligkeitsdatum für die New Hampshire Interest \&... und Business Tax-Returns am Dienstag, 18. April 2017 fällig sein. Die Fälligkeitsdaten für alle anderen Steuerarten mit einem Fälligkeitsdatum vom 15. April sind nicht von... } \\\cmidrule(lr){2-2} \cmidrule(lr){3-3}

\multirow{1}{*}{\textbf{Question}} & \parbox[t]{6cm}{ n Geschäftssteuer fällig } & { nh Geschäftssteuer Fälligkeitsdatum beeinflusst. }\\\cmidrule(lr){2-2}\cmidrule(lr){3-3}
\multirow{1}{*}{\textbf{LLM Score}} & \makecell[c]{\textbf{2}} & \makecell[c]{\textbf{3}} \\

\bottomrule[1.5pt]

\end{tabular}
\caption{Samples of translation result using the ``Separate'' and ``IT+RP'' method for the QG dataset.}
\label{tab:qg}
\end{table*}

Table~\ref{tab:qg} illustrates examples of results obtained from various translation methods applied to the QG dataset. As evident from the examples, translations that consider relation exhibit greater fidelity in preserving the content, sentence structure, and similarity of the passage and question to the reference.

\begin{figure*}[t]
\centering
\includegraphics[width=0.8\linewidth]{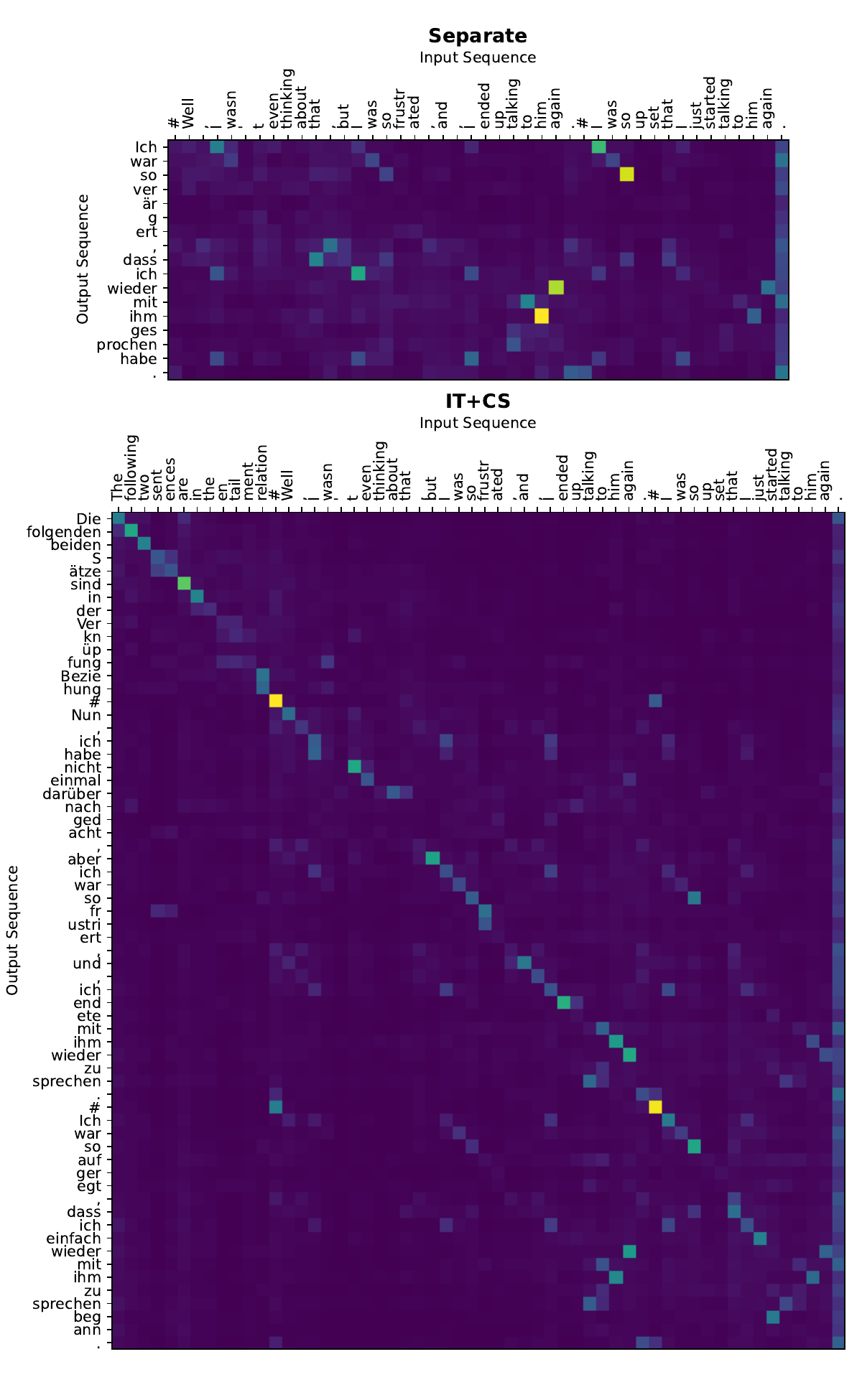}
 \caption{Cross-attention map in translating NLI data via NLLB-600M.} 
 \label{fig:attention_nli}
\end{figure*}

\begin{figure*}[t]
\centering
\includegraphics[width=0.9\linewidth]{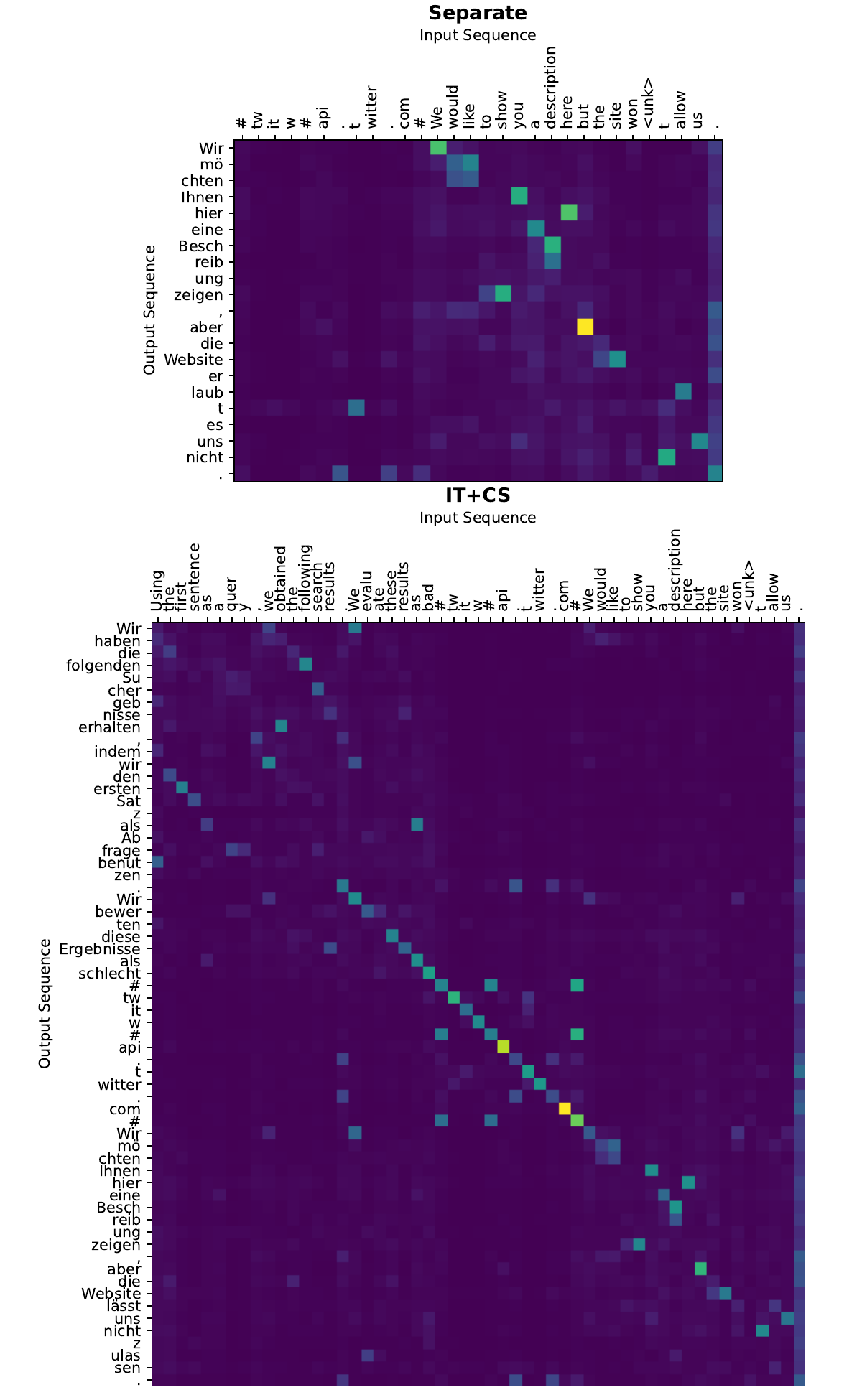}
 \caption{Cross-attention map in translating WPR data via NLLB-600M.} 
 \label{fig:attention_wpr}
\end{figure*}

\begin{figure*}[t]
\centering
\includegraphics[width=0.9\linewidth]{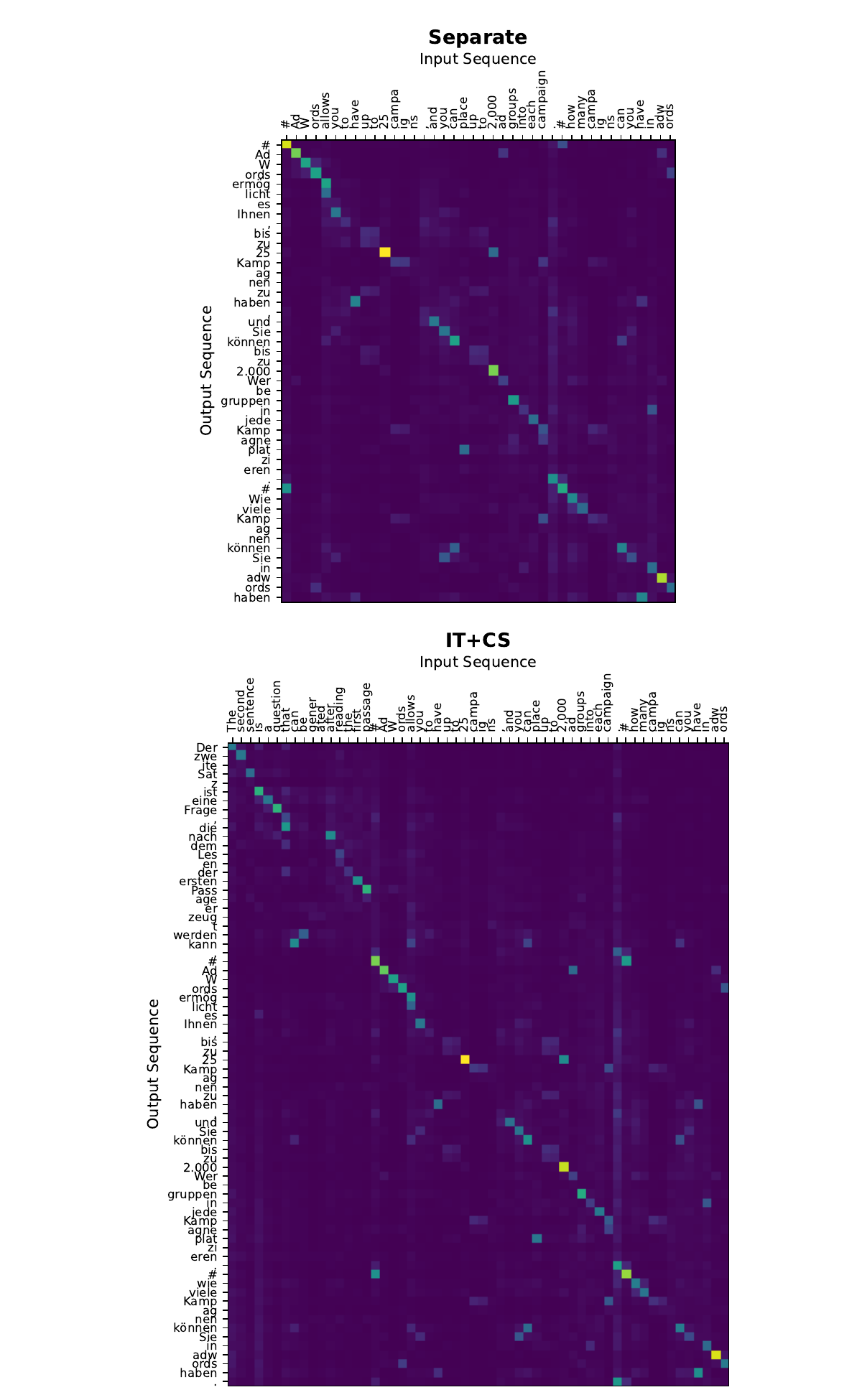}
 \caption{Cross-attention map in translating QG data via NLLB-600M.} 
 \label{fig:attention_qg}
\end{figure*}

Additionally, the results of the attention map analysis are depicted in Figure~\ref{fig:attention_nli}, Figure~\ref{fig:attention_wpr}, Figure~\ref{fig:attention_qg}. As can be observed from the results, when translation is conducted using the No CS approach, it is possible to refer to the context of each other, however, it frequently leads to the loss of the given IT. Conversely, when translating multifaceted data via the IT+CS method, semantic interreference between data components is observed, leading to superior translation quality and effective preservation of IT.

\label{app:qualitive}

\begin{table*}[ht]
\centering
\renewcommand\arraystretch{2} 
\setlength\extrarowheight{-2pt} 
\small
\begin{tabular}{c|l|l|l}
\toprule[1.5pt]
\multicolumn{1}{c}{} & \multicolumn{1}{c}{\textbf{NLI}} & \multicolumn{1}{c}{\textbf{WPR}} & \multicolumn{1}{c}{\textbf{QG}} \\ \cmidrule(lr){2-2}\cmidrule(lr){3-3}\cmidrule(lr){4-4}

\multirow{6}{*}{\textbf{Components}} & 
\parbox[t]{4cm}{
- \textbf{Premise}: One of our number will carry out your instructions minutely \\ 
- \textbf{Hypothesis}: A member of my team will execute your orders with immense precision . \\ 
- \textbf{Label}: Entailment} & 
\parbox[t]{4cm}{
- \textbf{Snippet}: you have chosen this item to be automatically replenished at the above selected frequency \\
- \textbf{query}: philosophy skin care \\
- \textbf{title}: philosophy.com - skin care, fragrance, perfume, bath and ... \\
- \textbf{label}: 4 (quality score)
} & \parbox[t]{4cm}{
- \textbf{Passage}: born on august 1 , 1779 , in frederick county , maryland , francis scott key became a lawyer who witnessed the british attack on fort mchenry during the war of 1812 .\\
- \textbf{Question}: when was francis scott key born
} \\ \cmidrule(lr){2-2}\cmidrule(lr){3-3}\cmidrule(lr){4-4}

\multirow{6}{*}{\makecell[c]{\textbf{No CS} \\ \textbf{(Only IT)}}} & 
\parbox[t]{4cm}{
\# One of our number will carry out your instructions minutely \\ 
\# A member of my team will execute your orders with immense precision .} & \parbox[t]{4cm}{
\# you have chosen this item to be automatically replenished at the above selected frequency \\
\# philosophy skin care \\
\# philosophy.com - skin care, fragrance, perfume, bath and ... \\
} & \parbox[t]{4cm}{
\# born on august 1 , 1779 , in frederick county , maryland , francis scott key became a lawyer who witnessed the british attack on fort mchenry during the war of 1812 .\\
\# when was francis scott key born
} \\ \cmidrule(lr){2-2}\cmidrule(lr){3-3}\cmidrule(lr){4-4}

\multirow{6}{*}{\textbf{Relation CS}} & 
\parbox[t]{4cm}{
The following two sentences are in the \textbf{entailment} relation \\
\# One of our number will carry out your instructions minutely \\ 
\# A member of my team will execute your orders with immense precision .} & \parbox[t]{4cm}{
Using the first sentence as a query, we obtained the following search results. We evaluate these results as \textbf{perfect} \\
\# you have chosen this item to be automatically replenished at the above selected frequency \\
\# philosophy skin care \\
\# philosophy.com - skin care, fragrance, perfume, bath and ... \\
} & \parbox[t]{4cm}{
The second sentence is a \textbf{question} that can be generated after reading the first \textbf{passage} \\
\# born on august 1 , 1779 , in frederick county , maryland , francis scott key became a lawyer who witnessed the british attack on fort mchenry during the war of 1812 .\\
\# when was francis scott key born
} \\

\bottomrule[1.5pt]
\end{tabular}
\caption{Sample translation sequences. We show examples of translation sequences utilizing "\#" as IT. For each task, we manually created label mapping to fit the original task objective. Specifically, we utilize [0: "entailment", 1: "neutral", 2: "contradiction"] for NER, and [4: "perfect", 3: "excellent", 2: "good", 1: "fair", 0: "bad"] for WPR.}
\label{tab:samples}
\end{table*}

\end{document}